\documentclass[11pt,letterpaper]{article}
\usepackage{emnlp2017}
\usepackage{times}
\usepackage{latexsym}
\usepackage{times}
\usepackage{latexsym}
\usepackage{amsfonts}
\usepackage{CJK}
\usepackage{amsmath}
\usepackage{bm}
\usepackage{graphicx}
\usepackage{url}
\usepackage{booktabs}
\usepackage{float}
\usepackage{multirow}
\usepackage{ulem}
\usepackage{xcolor} 

\emnlpfinalcopy

\def\emnlppaperid{***}

\title{A Constrained Sequence-to-Sequence \\Neural Model for Sentence Simplification}

\author{
	Yaoyuan Zhang$^{\dagger}$, Zhenxu Ye$^{\dagger}$, Yansong Feng, Dongyan Zhao, Rui Yan$^{*}$ \\
	Institute of Computer Science and Technology, Peking University, China \\
  {\tt $\{$zhang${\_}$yaoyuan, yezhenxu, fengyansong, zhaody,   ruiyan$\}$@pku.edu.cn}}

\date{}

\begin{document}

\maketitle

\begin{abstract}
	\renewcommand{\thefootnote}{$\dagger$}
	\footnotetext{Equal contribution.}
	Sentence simplification reduces semantic complexity to benefit people with language impairments. Previous simplification studies on the sentence level and word level have achieved promising results but also meet great challenges. For sentence-level studies, sentences after simplification are fluent but sometimes are not really simplified. For word-level studies, words are simplified but also have potential grammar errors due to different usages of words before and after simplification. In this paper, we propose a two-step simplification framework by combining both the word-level and the sentence-level simplifications, making use of their corresponding advantages. Based on the two-step framework, we implement a novel constrained neural generation model to simplify sentences given simplified words. The final results on Wikipedia and Simple Wikipedia aligned datasets indicate that our method yields better performance than various baselines.
\end{abstract}

\section{Introduction}

\textit{Sentence simplification} is a standard NLP task of reducing reading complexity for people who have limited linguistic skills to read. In particular, children, non-native speakers and individuals with language impairments such as Dyslexia \cite{Rello}, Aphasic \cite{Carroll} and Autistic \cite{Evans}, would benefit from the task which makes sentences easier to understand. There are two categories for the task: lexical simplification and sentence simplification. Both categories enable a paraphrasing process, turning a normal input into a simpler output while maintaining the same/similar semantics between the input and the output.

Inspired by the great achievements of machine translation, several studies treate the sentence simplification problem as monolingual translation task and achieve promising results \cite{Specia, Zhu, Coster, Wubben, Xu}. These studies apply the phrase-based statistical machine translation (PB-SMT) model or the syntactic-based translation model (SB-SMT). Both PB-SMT and SB-SMT require high-level features or even rules empirically chosen by humans. Recently, neural machine translation (NMT) based on sequence-to-sequence model (seq2seq) \cite{Cho, Sutskever, Bahdanau} shows more powerful capabilities than traditional SMT systems. NMT applies deep learning regimes and extracts features automatically without human supervision.

We observe that sentence simplification usually means the simple output sentence is derived from the normal input sentence with parts of terms changed, as shown in Table \ref{tab:one}. Due to such an intrinsic character, applying machine translation methods directly is likely to generate a completely identical sentence as the input sentence, no matter standard SMT or NMT. Although MT methods indeed have advanced the research of sentence simplification tasks, there is still plenty of room for improvements. To the best of our knowledge, there are few competitive simplification models using translation models even with neural network architectures so far.

Besides sentence-level simplification, there is lexical simplification which substitutes long and infrequent words with their shorter and more frequent synonyms by complex word identification, substitution generation, substitution selection and other processes. Recent lexical simplification models by Horn et al. \shortcite{Horn}, Glava\v s et al. \shortcite{Glavas}, Paetzold et al. \shortcite{Paetzold} and Pavlick et al. \shortcite{Pavlick} have accumulated substantial numbers of synonymous word pairs. This makes it possible for us to simplify complex words before simplifying the whole sentence. However, even though synonyms have similar semantic meaning, they might have different usages. Replacing complex words with their simpler synonyms is an intuitive way to simplify sentences, but not always works due to potential grammatical errors after the switchings (shown in Table \ref{tab:one}). Moreover, lexical substitution is just one way to simplify sentences. We can also simplify sentences by splitting, deletion, reordering and so on.

For the sentence-level simplification, we generally obtain output results with few grammar errors although it does not guarantee that they are simplified; for the lexical-level simplification, we can simplify the complicated parts of the sentences but it does not always guarantee that they are grammatically fluent. It is an intuitive and exciting idea to combine both methods together and make use of their corresponding advantages so that we can obtain simplified sentences with good readability and fluency. To be more specific, the simplification process of an input sentence is conducted in two steps. 1) We first identify complex word(s) and replace them with their simpler synonyms according to a pre-constructed knowledge base\footnote{A knowledge base such as PPDB contains millions of paraphrasing word pairs to change between simple words and complex words \cite{Pavlick}}. 2) The second step is to generate a legitimate sentence given the simplified word(s) with appropriate syntactic structures and grammar. Another key issue for the second step is that we need to maintain the same/similar semantic meaning of the input sentence. To this end, we still stick to the translation paradigm by translating the complex sentence into a simple sentence.

In this paper, our contributions are as follows:

$\bullet$ We propose a two-step simplification framework to combine the advantages of both word-level simplification and sentence-level simplification to make the generated sentences fluent, readable and simplified.

$\bullet$ We implement a novel constrained seq2seq model which fits our task scenario: certain word(s) are required to exist in the seq2seq process. We start from the constraint of one given word and extend the constraints to multiple given words. 

We evaluate the proposed method and neural model on English Wikipedia and Simple English Wikipedia datasets. The experimental results indicate that our model achieves better results than a series of baseline algorithms in terms of iBLEU scores, Flesch readability and human judgments. This paper is organized as follows. In Section \ref{sect:second} we review related works, and describe our proposed method and model in Section \ref{sect:third}. In Section \ref{sect:fouth}, we describe the experimental setups and show results. In Section \ref{sect:fifth}, we conclude our paper and discuss the future work.

\begin{table}
	\renewcommand{\multirowsetup}{\centering}  
	\begin{tabular}{c|p{4cm}}
		\hline
		\toprule 
		\hline
		\bf \multirow{5}{*}{Normal Sentence} & In  {\bf the last decades of} his life, dukas became well known as a teacher of {\bf composition}, with many famous students.\\
		\hline
		\bf \multirow{4}{*}{Simple Sentence} & {\bf Later} in his life, dukas became well known as a {\bf music} teacher, with many famous students. \\
		\hline
	\end{tabular}
	\caption{When ``{\bf the last decades of}'' is replaced with the simplified word ``{\bf later}'', the words ``{\bf later}'' and ``{\bf in}'' need to be reordered to guarantee the correct grammaticality of the output sentence. The word pair ``{\bf composition}'' and ``{\bf music}'' has the same situation as ``{\bf the last decades of}'' and ``{\bf later}''.}\label{tab:one}
\end{table}

\section{Related Work}
\label{sect:second}

In previous studies, researchers of sentence-level simplification mostly address the simplification task as a monolingual machine translation problem. Specia et al. \shortcite{Specia} use the standard PB-SMT implemented in Moses toolkit \cite{Koehn} to translate the original sentences to the simplified ones. Similarly, Coster and Kauchak \shortcite{Coster} extend the PB-SMT model by adding phrase deletion. Wubben et al. \shortcite{Wubben} make a further effort by reranking the Moses' n-best output based on their dissimilarity to the input. Most recently, Xu et al. \shortcite{Xu} have proposed a SB-SMT model, achieving better performance than Wubben's system. In general, sentence-level simplification maintains the semantic meaning and fluency but does not always guarantee the literal simplification.

As for word-level simplification, there are impressive results as well. Horn et al. \shortcite{Horn} extract over 30,000 paraphrase rules for lexical simplification by identifying aligned words in English Wikipedia and Simple English Wikipedia. Glava\v s et al. \shortcite{Glavas} employ GloVe \cite{Pennington} to generate synonyms for the complex words. Instead of using the parallel datasets, their approach only requires a single corpus. Paetzold et al. \shortcite{Paetzold} propose a new word embeddings model to deal with the limitation that the traditional models do not accommodate ambiguous lexical semantics. Pavlick et al. \shortcite {Pavlick} release about 4,500,000 simple paraphrase rules by extracting normal paraphrases rules from a bilingual corpus and reranking the simplicity scores of these rules by a supervised model. Thanks to their efforts, there is a large number of effective methods for identifying complex words, finding corresponding simple synonyms and selecting qualified substitutions. However, sometimes simplifying complicated words directly with simple synonyms violates grammar rules and usages.

Recent progress in deep learning with neural networks brings great opportunities for the development of stronger NLP systems such as neural machine translation (NMT). Deep learning is heavily driven by data, requiring few human efforts to create high-level features. Specifically, the sequence-to-sequence RNN model \cite{Cho, Sutskever, Bahdanau, Mou1, Mou2} has a remarkable ability to characterize word sequences and generate natural language sentences. However, the seq2seq NMT model still shares the problem with other MT-based methods: lack of literal simplification from time to time.

Overall, sentence-level and word-level sentence simplification both have their strengths and weaknesses. In this paper, we propose a two-step method fusing their corresponding advantages. 

\section{Neural Generation Model}
\label{sect:third}
We generate simplified sentences using a sequence-to-sequence model trained on a parallel corpus, namely English Wikipedia and Simple English Wikipedia. We have simplified word(s) identified from the first step, but the standard sequence-to-sequence model cannot guarantee the existence of such word(s). Therefore, we propose a constrained sequence-to-sequence (Constrained Seq2Seq) model with the given simplified word(s) as constraints during the sentence generation.

\subsection{Methodology}
\label{ssect:third-one}
Since there has been many efforts working on the establishment of word simplification pairs \cite{Horn, Glavas, Paetzold, Pavlick}, we do not focus on the identification of words that require simplification or the methods of selecting what simpler words to switch. Instead, we change words according to these knowpledge base and proceed to the neural sentence generation model, assuming the word substitutions are correct based on previous studies on synonym alignments. To be more specific, given an input sentence, the simplification process is conducted in two steps:

$\bullet$ \textbf{Step 1.} According to previous studies on lexical substitution, we first identify complex words in the input sentence and then substitute them with their simpler synonyms;

$\bullet$ \textbf{Step 2.} Given the simplified words from the first step as constraints, we propose a constrained seq2seq model which encodes the input sentence as a vector (encoder) and decodes the vector into a simplified sentence (decoder). The generation process is conditioned on the simplified word(s) and consists of both backward and forward generation.

We proceed to introduce the proposed constrained seq2seq model in the next section.

\subsection{Constrained Seq2Seq Model}
\label{ssect:third-two}
Given an input sequence $\mathbf x=(x_1,\ldots, x_n), x\in\mathbb{R}^{V}$ and a switching word pair of a complex word $\bm x_c$ and its simpler synonym \textcolor[rgb]{0.980392, 0.501961, 0.447059}{$\bm y_s$}, we aim to generate a simplified sentence $\mathbf y=(y_1,\ldots, y_m), y\in\mathbb{R}^{V}$ as the output where \textcolor[rgb]{0.980392, 0.501961, 0.447059}{$\bm y_s$} is contained in $\mathbf y$, i.e., \textcolor[rgb]{0.980392, 0.501961, 0.447059}{$\bm y_s$} $\in \mathbf y$. There could be multiple constraint words and we start from the simplest situation with only one constraint word. Here $n$ and $m$ denote the lengths of the source and target sentences respectively, $V$ is the vocabulary size of the source and target sentences.

The simpler word \textcolor[rgb]{0.980392, 0.501961, 0.447059}{$\bm y_s$} splits the output sentence into two sequences: \textit{backward sequence} $\mathbf y_ \mathbf b$ $=(y_{s-1},\ldots, y_1)$ and \textit{forward sentence} $\mathbf y_ \mathbf f$ $=(y_{s+1},\ldots, y_m)$.
The joint probabilities of $\mathbf y_\mathbf b$ and $\mathbf y_ \mathbf f$ are:
\begin{equation}
\begin{aligned}
\nonumber& p(\mathbf y_ \mathbf b) =\prod_{i=1}^{s-1}p(y_{s-i} | y_s,\ldots, y_{s-i+1},\mathbf x)\\
\end{aligned}
\end{equation}
\begin{equation}
\begin{aligned}
& p(\mathbf y_ \mathbf f) =\prod_{i=1}^{m-s}p(y_{s+i} | y_1,\ldots, y_s,\ldots,y_{s+i-1},\mathbf x)\\
\end{aligned}
\end{equation}

\begin{figure}
	\centering
	\includegraphics[height=4.5cm,width=7.5cm]{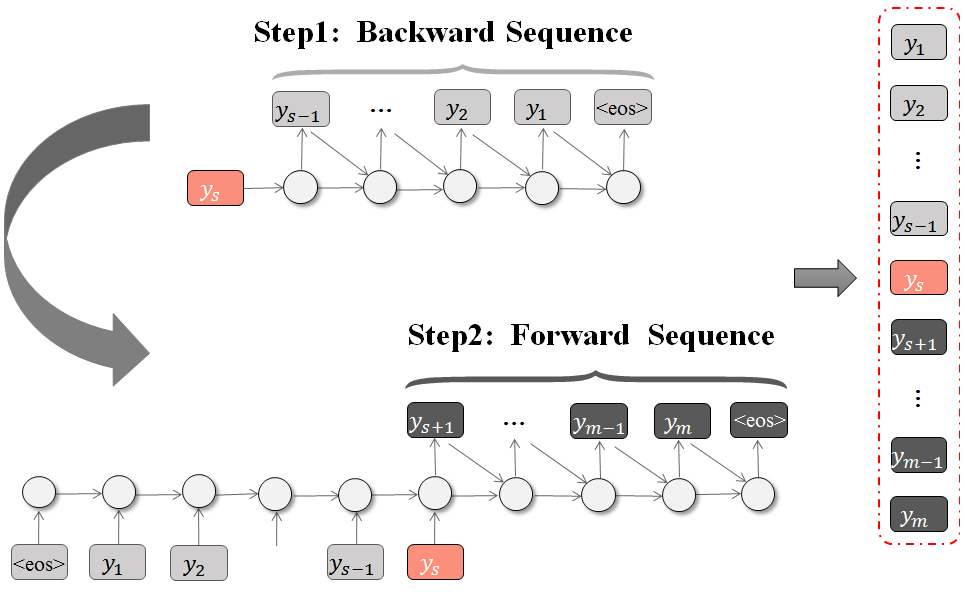}
	\caption{An overview of the backward and forward model that generates the simplified sentence where one target simplified word must appear. }\label{fig:one}
\end{figure}
\begin{figure*}[t]
	\centering
	\includegraphics[height=8.5cm,width=14.5cm]{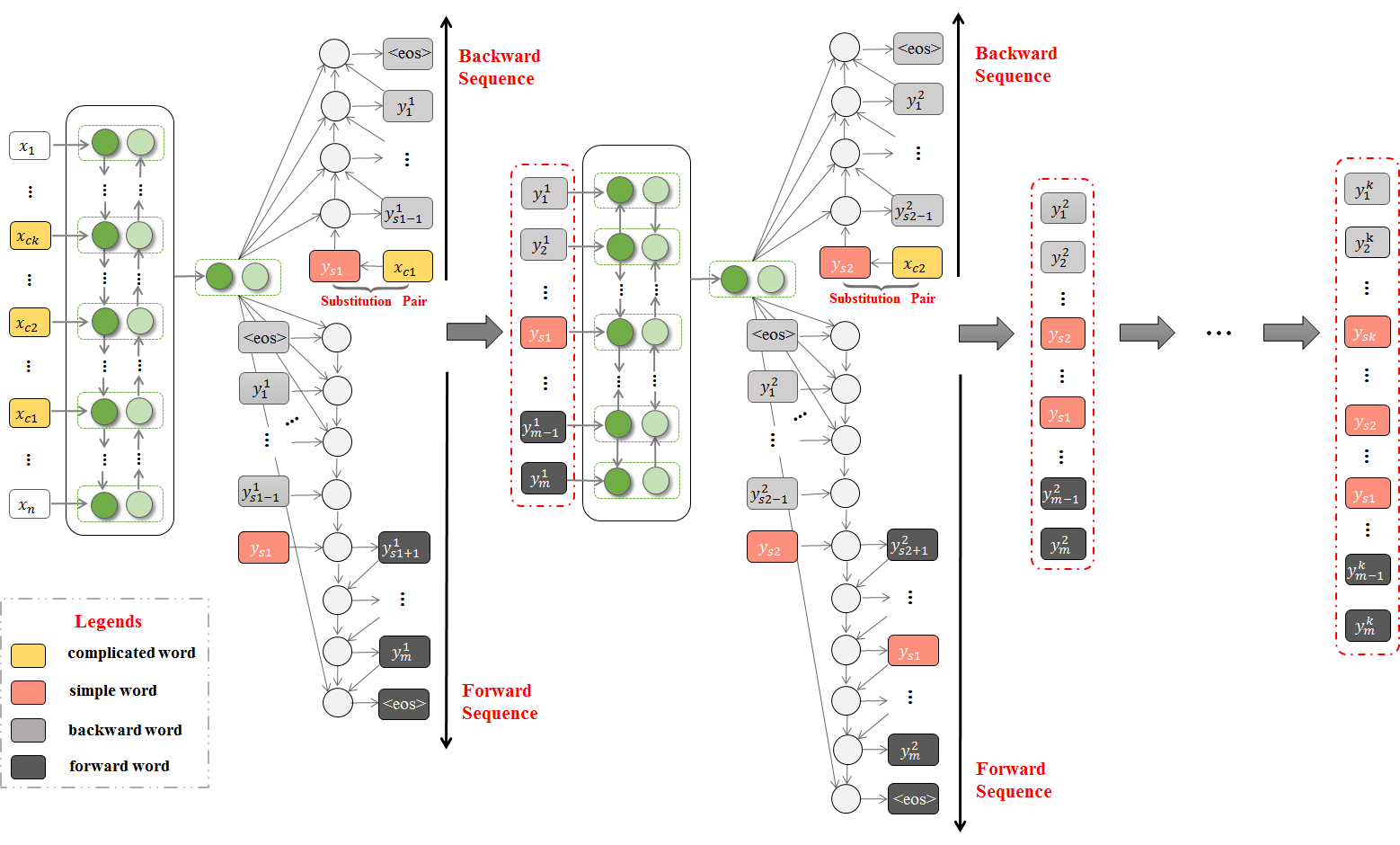}
	\caption{An overview of our system extended the constrained seq2seq model that generates simplified sentence where two target simplified words must appear.}\label{fig:two}
\end{figure*}

As shown in Figure \ref{fig:one}, to generate a sequence $\mathbf y$ with a constraint \textcolor[rgb]{0.980392, 0.501961, 0.447059}{$\bm y_s$}, we first generate a sequence from $y_s$ to $y_1$. Then, we generate a forward sequence $\mathbf y_ \mathbf f$ conditioned on the generated sequence from $y_1$ to $y_s$. In this way, the output sequence is $\mathbf y$ = $\{\mathbf y_ \mathbf b^{-1},$ \textcolor[rgb]{0.980392, 0.501961, 0.447059}{$\bm y_s$}, $\mathbf y_ \mathbf f\}$, where $\mathbf y_ \mathbf b^{-1}$ is the reverse of $\mathbf y_ \mathbf b$. In our paper, we apply the bi-directional recurrent neural network (BiRNN) \cite{Schuster} with gated recurrent units (GRUs) \cite{Cho} for both the backward and forward generation process. We encode the input sequence as follows:
\begin{equation}
\begin{aligned}
& \bm{z}_t = \sigma(W_z\bm{e}_t  + U_z\overrightarrow{\bm{h}}_{t-1})\\
& \bm{r}_t = \sigma(W_r\bm{e}_t  + U_r\overrightarrow{\bm{h}}_{t-1})\\
& \tilde{\bm{h}}_t = tanh( W_h \bm{e}_t + U_h [\bm{r_t} \circ \overrightarrow{\bm{h}}_{t-1}]) \\
& \overrightarrow{\bm{h}}_t = (1 - \bm{z}_t) \circ \bm{\overrightarrow{h}_{t-1}} + \bm{z}_t \circ \tilde{\bm{h}}_{t}\\
\end{aligned}
\end{equation}
where $\bm{e}_{t}\in\mathbb{R}^{E}$ is the embedding vector of word $x_t$ and $E$ denotes the word embedding dimensionality; $W_z, W_r, W_h\in \mathbb{R}^{dim \times E}$, $U_z, U_r, U_h\in \mathbb{R}^{dim \times dim}$ are weight matrices\footnote{Bias term are omitted for simplicity} and $dim$ denotes the number of hidden states; $\overrightarrow{\bm{h}}_t\in\mathbb{R}^{dim}$ is the hidden state of BiRNN's forward direction at time $\bm t$. Likewise, the hidden state of BiRNN's backward direction is denoted as $\overleftarrow {\bm{h}}_t$.

As last, we concatenate the bidirectional hidden states as the sentence embedding:
\begin{equation}
\bm h_t = \biggl[ \begin{array}{ll}
\overrightarrow {\bm{h}}_t^\top; \overleftarrow {\bm{h}}_t^\top
\end{array} \biggl]^\top
\end{equation}

{\bf RNN Decoder} We adapt the gated recurrent unit with the attention mechanism at the decoder part, where the hidden state $s_t \in\mathbb{R}^{dim}$ is computed by:
\begin{equation}
\begin{aligned}
\bm{z}^{'}_t & =\sigma(W^{'}_z\bm{e}^{'}_{t-1} + U^{'}_z{\bm{s}}_{t-1} + C_z\bm c_t )\\
\bm{r}^{'}_t & =\sigma(W^{'}_r\bm{e}^{'}_{t-1} + U^{'}_r{\bm{s}}_{t-1} + C_r\bm c_t )\\
\tilde{\bm s}_t &= tanh( W_s \bm{e}^{'}_{t-1} + U^{'}_s [\bm{r}^{'}_t \circ {\bm{s}}_{t-1}] + C_s\bm c_t ) \\
{\bm{s}}_t &= (1 - \bm{z}^{'}_t) \circ \bm{s}_{t-1} + \bm{z}^{'}_t \circ \tilde{\bm{s}}_{t}\\
\end{aligned}
\end{equation}
where $\bm{e}^{'}_{t-1}\in\mathbb{R}^{E}$ is the embedding vector of word $y_t$ and $E$ denotes the word embedding dimensionality; $W^{'}_z, W^{'}_r, W_s\in \mathbb{R}^{dim \times E}$, $U^{'}_z, U^{'}_r, U^{'}_s\in \mathbb{R}^{dim \times dim}$, $C_z, C_r, C_s\in \mathbb{R}^{dim \times 2dim}$ are weight matrices and $dim$ denotes the number of hidden states. The initial hidden state $s_0$ is computed by $s_0 = tanh(W_{sh}H_{mean})$, where $H_{mean}$ is the mean value of all hidden states of {\bf Encoder} and $W_{sh}\in \mathbb{R}^{dim \times dim}$.

The context vector $c_t$ is recomputed at each step by an alignment model:
\begin{equation}
\begin{aligned}
c_t & = \sum_{j=1}^{n}\alpha_{tj}h_j \\
\alpha_{tj} & = {\frac{exp(e_{tj})}{\sum _{k=1}^{n} exp(e_{tk})}} \\ 
\end{aligned}
\end{equation}
\begin{equation}
\begin{aligned}
\nonumber e_{tj} & = a( s_{t-1},h_j)
\end{aligned}
\end{equation}
where $\alpha_{tj}$ is the alignment weight implemented by $a$ function which is an attention mechanism to align the input token $x_j$ and the output token $y_t$. The more tightly they match each other, the higher scores they obtain.

With the decoder state $s_t$ and the context vector $c_t$, we approximately compute each conditional probability either in the backward sequence $\mathbf y_ \mathbf b$ or the forward sequence $\mathbf y_ \mathbf f$ as Eq.$(7)$:
\begin{equation}
\begin{aligned}
p(y_t | y_1,\ldots, y_{t-1},\mathbf x) & = p(y_t | y_{t-1}, s_t, c_t)
\end{aligned}
\end{equation}

According to the Eq.$(1)$ and Eq.$(2)$, we in turn obtain the backward sentence $\mathbf y_\mathbf b$ = $(y_{s-1},\ldots, y_1)$ and the forward sentence $\mathbf y_\mathbf f$ = $(y_{s+1},\ldots, y_m)$, with the maximal estimated probability by beam search. Finally, we concatenate the reverse backward sequence $\mathbf y_\mathbf b$, the simpler word \textcolor[rgb]{0.980392, 0.501961, 0.447059}{$\bm y_s$} and the forward sequence $\mathbf y_\mathbf f$ to output the entire sentence $\mathbf y = (y_1, \ldots, $\textcolor[rgb]{0.980392, 0.501961, 0.447059}{$\bm y_s$}$, \ldots, y_m)$. Notice that \textcolor[rgb]{0.980392, 0.501961, 0.447059}{$\bm y_s$} can exist at any position in $\mathbf y$.
%be any token of the output sentence, where the position index ranges from $1$ to $m$.

\subsection{Multi-constrained Seq2Seq}
\label{ssect:third-three}
We just illustrated how to put a single constraint word into the sequence-to-sequence generation process, while actually there can be more than one constraint words which are simplified before the sentence generation, as shown in Table 1. We extend the single constraint into multiple constraints by a Multi-Constrained Seq2Seq model. Without loss of generosity, we define the multiple keywords as \textcolor[rgb]{0.980392, 0.501961, 0.447059}{$\bm y_{s1}, \bm y_{s2}$}$, \ldots,$ \textcolor[rgb]{0.980392, 0.501961, 0.447059}{$\bm y_{sk}$} and illustrate Multi-Constrained Seq2Seq in Figure \ref{fig:two}.% describes the overall system.

%In the section, we extend the seq2BF model to handle the situation where multiple separate words (phrases) must appear in the simple output sentence. Here we

We first illustrate the situation with two simplified words, i.e., $k$ = $2$, namely \textcolor[rgb]{0.980392, 0.501961, 0.447059}{$\bm y_{s1}$} and \textcolor[rgb]{0.980392, 0.501961, 0.447059}{$\bm y_{s2}$}. We generally take the complex word with the least term frequency as the first constrained word \textcolor[rgb]{0.980392, 0.501961, 0.447059}{$\bm y_{s1}$} and use the same method as in Section \ref{ssect:third-two} to generate the first output sentence $\mathbf y^{1}=(y^{1}_1, \ldots, $ \textcolor[rgb]{0.980392, 0.501961, 0.447059}{$\bm y_{s1}$}$, \ldots, y^{1}_m)$. In the second round of generation, we take the first output sentence $\mathbf y^{1}$ as the input and generate the second output sentence $\mathbf y^{2}$ with \textcolor[rgb]{0.980392, 0.501961, 0.447059}{$\bm y_{s2}$} as the constraint. Compared with the single-pass generation with a single constraint word, we have an output sentence $\mathbf y^{2}=(y^{2}_1, \ldots, $\textcolor[rgb]{0.980392, 0.501961, 0.447059}{$\bm y_{s2}$}$, \ldots,$ \textcolor[rgb]{0.980392, 0.501961, 0.447059}{$\bm y_{s1}$}, \ldots, $y^{2}_m)$ with two constraint words after a two-pass generation. The relative position of \textcolor[rgb]{0.980392, 0.501961, 0.447059}{$\bm y_{s1}$} and \textcolor[rgb]{0.980392, 0.501961, 0.447059}{$\bm y_{s2}$} depends on the input sentence .

%After two rounds of generation, we get the final output sentence .
When there are more than two constraint words, i.e., $k > 2$, the system architecture remains the same with more repeating passes of generation included (shown in Figure \ref{fig:two}). To decode the $k$-th output sentence $\mathbf y^{k}$, we encode the output sentence $\mathbf y^{k-1}$ to the next word embeddings of the multi-constrained seq2seq model. 

Note that after each pass of generation, other constrained words may be deleted or simplified already. if there are complex word(s) which need to be simplified, the system will repeat the simplification process.
%We observe that more than 90\% of the simplification cases are covered by one and two constraint words.

\section{Experiment}
\label{sect:fouth}
\subsection{Dataset and Setups}
\label{ssect:fouth-one}
We evaluate our proposed approach on the parallel corpus from Wikipedia and Simple Wikipedia in English\footnote{This dataset is available at \url{http://www.cs.pomona.edu/~dkauchak/simplification}}. We randomly split the corpus into 123,626 sentence pairs (each pair as a normal sentence and its simplification in parallel) for training, 5,000 sentences for validation and 600 sentences for testing. There can be noises in the dataset. We filter out test samples when the output is identical as the input without any simplification. We also applied lowercasing preprocess to all samples. Our vocabulary size is 60,000 while out-of-vocabulary words are mapped to the token ``unk''.

The RNN encoder and decoder of our model both have 1,000 hidden units; the word embedding dimensionality is 620. We use the Adadelta \cite{Zeiler} to optimize all parameters.

\subsection{Comparison Methods}
\label{ssect:fouth-two}
In this paper, we conduct the experiments on the English Wikipedia and Simple English Wikipedia datasets to compare our proposed method against several representative algorithms. 

%compared four different approaches on the sentence simplification task with our models in this paper:

\setlength{\parindent}{0pt}\textbf{Moses}. It is a standard phrase-based machine translation model \cite{Koehn}.　

\setlength{\parindent}{0pt}\textbf{SBMT}. SBMT is a syntactic-based machine translation model \cite{Xu}, which is implemented on the open-source Joshua toolkit \cite{Post}. The simplification model is optimized to the SARI metric and leverages the PPDB dataset \cite{Pavlick} as a rich source of simplification operations.

\setlength{\parindent}{0pt}\textbf{Lexical Substitution}. This method only substitutes the complex words with the simplified word(s) which we use as the constraint word(s) in our model and leave other words of the input sentence unchanged. This model shares the same hypothesis as our model.

%based on the same hypothesis as ours is used for analysing the hypothesis' impact on our model.

\textbf{Seq2Seq}. The sequence-to-sequence model is the state-of-the-art neural machine translation model \cite{Cho} with the attention mechanism applied \cite{Bahdanau}. %The encoder and decoder apply attentional RNNs like us. It is tuned on the same training dataset and validation dataset as those in our model.

\textbf{Constrained Seq2Seq}. We propose a novel neural sentence generation model based on sequence-to-sequence paradigm with one constraint word. We use \textbf{Multi-Constrained Seq2Seq} to denote the scenario when there is more than one constraint words.

\subsection{Evaluation Metrics}
\label{ssect:fouth-three}
\begin{table*}[t]
	\centering
	\renewcommand\arraystretch{1.1}
	\begin{tabular}{c|c|c|c|c|c}
		\hline
		\toprule
		\hline
		& FK &BLEU(O, R) &BLEU(O, I) & iBLEU & SARI\\
		\hline
		\bf {English Wikipedia} & 13.32 &  28.19 & 100.0 & 15.37& 13.59\\
		\hline
		\bf {Simple English  Wikipedia} & 10.75 & 100.0 & 30.41 & 86.96&91.47
		\\
		\hline
		\hline
		{\bf Moses} &13.31 & 28.28 & 99.62 &15.49 &14.87\\
		\hline
		{\bf SBMT}& 13.15 & 27.81 & 97.07&15.33 &20.62\\
		\hline
		{\bf Seq2Seq} &12.74 & 25.51 &  66.94 &  16.27  &33.16\\
		\hline
		{\bf Lexical Substitution} &13.34  &\bf 30.44 & 84.31 &18.97 & \bf 46.29\\
		\hline
		{\bf Constrained Seq2Seq} &\bf 11.33 &\bf 29.44 & 62.30  & \bf 20.26 & 43.44\\
		\hline
		{\bf Multi-Constrained Seq2Seq} &\bf 11.02 & 27.94 & 52.72 & \bf 19.87& \bf 43.98\\
		\hline
		
		\hline
	\end{tabular}
	\caption{Automatic evaluation of several simplification systems.}\label{tab:two}
\end{table*}
\begin{table*}[t]
	\centering
	\renewcommand\arraystretch{1.1}
	\begin{tabular}{c|c|c|c|c}
		\hline
		\toprule
		\hline
		& Grammaticality & Meaning & Simplicity & Same Percentage\\
		\hline
		\bf {English Wikipedia} & 4.00 & 4.00 & 0.00 & 120/120 \\
		\hline
		\bf {Simple English  Wikipedia} &3.55& 2.83 & 2.20 & 0/120\\
		\hline
		\hline
		{\bf Moses} & 3.99 & 3.99 & 0.02 & 116/120\\
		\hline
		{\bf SBMT}& 3.97 & 3.94 & 0.19 & 99/120 \\
		\hline
		{\bf Lexical Substitution} & 3.01 & 3.36 & 1.24& 2/120\\
		\hline	
		{\bf Seq2Seq} & 3.28 & 3.45 &  0.91 & 30/120\\
		\hline
		{\bf Constrained Seq2Seq} & 3.16 & 2.81 & 1.50 & 0/120 \\
		\hline
		{\bf Multi-Constrained Seq2Seq} & 2.60 & 2.65 & 1.62 & 0/120 \\
		\hline
		\hline
	\end{tabular}
	\caption{Human evaluation of several simplification systems dicussed in our paper.}\label{tab:three}
\end{table*}
\ULforem
\begin{table*}[t]
	\centering
	\renewcommand\arraystretch{1.1}
	\begin{tabular}{c|p{10cm}}
		\hline
		\toprule
		\hline
		\bf \multirow{3}{*}{English Wikipedia} & parkes became a key country location after the completion of the railway in 1893 , serving as a hub for a great deal of passenger and freight transport until the 1980s . \\
		\hline
		\bf \multirow{3}{*}{Simple English  Wikipedia} & parkes was an {\bf important} transport {\bf center} after the railway was built in 1893 . {\bf many} passenger and freight trains stopped at parkes up until the 1980s . \\
		\hline
		\hline
		\multirow{3}{*}{\bf Moses} & parkes became a key country location after the completion of the railway in 1893 , serving as a hub for a great deal of passenger and freight transport until the 1980s .
		\\
		\hline
		\multirow{3}{*}{\bf SBMT} & parkes became a key country location after the completion of the railway in 1893 , serving as a hub for a great deal of passenger and freight transport until the 1980s .
		\\
		\hline
		\multirow{3}{*}{\bf Seq2Seq}& parkes became a key country location after the completion of the railway in 1893 , serving as a hub for a great deal of passenger and freight transport until the 1980s .
		\\
		\hline
		\multirow{3}{*}{\bf Lexical Substitution} &  parkes became a  {\bf important} country location after the completion of the railway in 1893 , serving as a {\bf center} for {\bf many} passenger and transport the 1980s .  \\
		\hline
		\multirow{3}{*}{\bf Constrained Seq2Seq} & parkes became a key country location after the completion of the railway in 1893 , serving as a {\bf center} of \sout{a great deal of} passenger and freight transport until the 1980s . \\
		\hline
		\multirow{3}{*}{\bf Multi-Constrained Seq2Seq} & parkes became {\bf an important} country location after the completion of the railway in 1893. {\bf it became a center} of \sout{a great deal of} passenger and freight transport until the 1980s .  \\
		\hline
		\hline
	\end{tabular}
	\caption{Cases Study}\label{tab:four}\vspace{-2mm}
\end{table*}
\normalem

\textbf{Automatic Evaluation} To evaluate the performance of different methods for the simplification task, we leverage four automatic evaluation metrics\footnote{The highest n-gram order of all correlation related metrics is set to 4 in our experiments}: Flesch-Kincaid grade level (FK) \cite{Kincaid},  SARI \cite{Xu}, BLEU \cite{Papineni} and iBLEU \cite{Sun}. FK is widely used for readability. SARI evaluates the simplicity by explicitly measuring the goodness of words that are added, deleted and kept. BLEU is originally designed for MT and evaluates the output by using n-gram matching between the output and the reference. Several studies indicate that BLEU alone is not really suitable for the simplification task \cite{Zhu, Stajner2, Xu}. In many cases of sentence simplification, the output sequence looks similar to the input sequence with only part of it simplified. Due to this situation, there is prominent insufficiency for the standard BLEU metric: even though the output sequence does not perform any simplification operations on the input sequence, it is still likely to obtain a high BLEU score. It is necessary to penalize the output sentence that is too similar to the input sentence. Therefore, the iBLEU metric is more suitable for simplification as it balances similarity and simplicity. Given an output sentence $O$, a reference sentence $R$ and input sentence $I$, iBLEU\footnote{$\alpha$ is set to 0.9 as suggested by Sun et al. \shortcite{Sun}} is defined as:
\vspace{-1mm}
\begin{equation}
\begin{aligned}
iBLEU = \alpha \times BLEU(O, R)-\\(1-\alpha) \times BLEU(O, I)
\end{aligned}
\end{equation}

\vspace{-1mm}
\textbf{Human Evaluation} 
Human judgment is the ultimate evaluation metric for all natural language processing tasks. We randomly select 120 source sentences from our test dataset and invite 20 graduate students (include native speakers) to evaluate the simplified sentences by all systems according to the source sentence. For fairness, we conduct a blind review: the evaluators are not aware which methods produce the simplification results. Following earlier studies \cite{Wubben, Xu}, we asked participants to rate {\em Grammaticality} (the extent to which the simplified sentence is grammatically correct and fluently readable), {\em Meaning} (the extent to which the simplified sentence has the same meaning as the input) and {\em Simplicity} (the extent to which the simplified sequence maintains similar meaning). All these three human evaluation metrics are in 5-points scale from lowest 0 points to highest 4 points. Note that if one generated sentence is identical to the source sentence, we rate {\em Grammaticality} with 4 points, {\em Meaning} with 4 points and {\em Simplicity} with 0 points for this target sentence.

\subsection{Overall Performance}
\label{ssect:fouth-four}
The automatic evaluation results are listed in Table \ref{tab:two}. Moses has the worst performance. It obtains a fair BLEU(O, R) score that is 28.28. But its BLEU(O, I) score is 99.62, indicating that Moses fails to simplify most of the sentences. As this failure, its FK, iBLEU and SARI scores are all quite low. SBMT has the similar performance like Moses and neither simplify the output sentences nor promote the readability. The overall results of the Seq2Seq system are better than Moses and SBMT. Though its BLEU(O, R) score is little lower than Moses and SBMT, its output sentences are not mostly identical to the input sentences as its BLEU(O, I) score is only 66.94. It also achieves better FK(12.74), iBLEU(16.27) and SARI(33.16) scores than Moses and SBMT. Lexical Substitution only substitutes the complex words so that it obtains the highest BLEU(O, R) and SARI scores. But it gets the worst FK readability. In general, both Constrained Seq2Seq and Multi-Constrained Seq2Seq under our proposed framework outperform baselines. They have higher similarities to the reference and lower similarities to the input than other systems. So the iBLEU scores of our two systems are higher than baselines, which are 20.26 and 19.87 respectively. The SARI score of our two systems is also pretty high. As for FK readability, our two systems achieve the best result. 

The human evaluation results are displayed in Table \ref{tab:three}. Moses generates 116 sentences that are completely identical to the input sentences. As the {\em Grammaticality} of the identical sentences are rated with 4 points, the {\em Meaning} with 4 points and the {\em Simplicity} with 0 points, Moses gets the highest score (3.99) both in {\em Grammaticality} and {\em Meaning} but obtains the lowest score (0.02) in {\em Simplicity}. Similar to Moses, SBMT generates 99 sentences that are not really simplified so that SBMT obtains similar results like Moses. Seq2Seq outperforms Moses and SBMT systems judged by the overall performance and obtains 3.28 in {\em Grammaticality}, 3.45 in {\em Meaning} and 0.96 in {\em Simplicity}. The results of Lexical Substitution in {\em Meaning} and {\em Simplicity} are rather high. But as shown by the score of {\em Grammaticality}, the sentences generated by Lexical Substitution contain many grammar errors which are not surprising. Our Constrained Seq2Seq and Multi-Constrained Seq2Seq outperform in {\em Simplicity} than baselines. The {\em Meaning} scores of our systems are 2.81 and 2.65. Simple English Wikipedia has a quite similar score, 2.83, which indicates that to some extent, both our systems and Simple English Wikipedia have a semantic loss when simplifying sentences. As for {\em Grammaticality}, Constrained Seq2Seq is better than Lexical Substitution. Multi-Constrained Seq2Seq performs worse than Constrained Seq2Seq in {\em Grammaticality} but better in {\em Simplicity}. 

Judged by the overall performances, Constrained Seq2Seq and Multi-Constrained Seq2Seq outperform other off-the-shelf sentence-level simplification methods as their generated sentences are literally simplified and legitimate.

\subsection{Analysis and Case Studies}
\label{ssect:fouth-five}
In Table \ref{tab:four}, we show some typical examples of all systems. Among them, Moses, SBMT and the Seq2Seq model generate a completely identical sentence to the input sentence as they do in most cases. Lexical Substitution paraphrases the complex words {\em ``key''}, {\em ``hub''} and {\em ``a great deal of''} with the simple words {\em ``important''}, {\em ``center''} and {\em ``many''}. As seen, the article for the word {\em important} should be changed from {\em ``a''} to {\em ``an''} but Lexical Substitution fails to deal with such kind of errors. As for our proposed model, it generates the output sentences conditioned on the simplified word {\em ``center''} and deletes the complex phrase {\em ``a great deal of''}. Taking the generated sentence of Constrained Seq2Seq model as a input, the Multi-Constrained Seq2Seq model substitutes the less frequent words {\em ``key''} with the word {\em ``important''}. It also changes the adverbial clause {\em ``serving as ... until the 1980s''} with a simple sentence structure {\em ``it became ... until the 1980s''}, which shows that our models are more flexible and more effective than other baseline systems. 

\section{Conclusion}
\label{sect:fifth}
In this paper, we propose a new two-step method for sentence simplification by combining word-level simplification and sentence-level simplification. We run experiments on the parallel datasets of Wikipedia and Simple Wikipedia and the results show that our methods outperform various baselines with better readability, flexibility and simplicity achieved. In the future, we plan to take more factors (e.g., sentence length or grammar rules) into account and formulate them as constraints into our proposed model.

\bibliography{emnlp2017}
\bibliographystyle{emnlp_natbib}
\end{document}